# Hand Biometrics in Digital Forensics

Asish Bera [*1], Debotosh Bhattacharjee[*2], Mita Nasipuri[*3]

[*]Department of Computer Science and Engineering, Jadavpur University, Kolkata-32, India.

asish.bera@gmail.com, debotosh@ieee.org, mitanasipuri@gmail.com

**Abstract.** Digital forensic is now an unavoidable part for securing the digital world from identity theft. Higher order of crimes, dealing with a massive database is really very challenging problem for any intelligent system. Biometric is a better solution to win over the problems encountered by digital forensics. Many biometric characteristics are playing their significant roles in forensics over the decades. The potential benefits and scope of hand based modes in forensics have been investigated with an illustration of hand geometry verification method. It can be applied when effective biometric evidences are properly unavailable; gloves are damaged, and dirt or any kind of liquid can minimize the accessibility and reliability of the fingerprint or palmprint. Due to the crisis of pure uniqueness of hand features for a very large database, it may be relevant for verification only. Some unimodal and multimodal hand based biometrics (e.g. hand geometry, palmprint and hand vein) with several feature extractions, database and verification methods have been discussed with 2D, 3D and infrared images.

**Keywords.** Forensics, fusion, hand biometrics, multibiometrics.

## 1 Introduction

Omnipresence requirement of security concerned applications in different domains is extremely indispensable and ineluctable in this digital age to protect the assets and properties from unauthorized access or identity theft. As the world population is rising day after day in higher magnitude and therefore the private information is increasing proportionately. The size of the database is becoming ultra large and its truly essential terabytes to store data in digitized versions. Studies said about 95% data have been stored in digital format through worldwide and more than 50% of them are not printed out. So, it is very difficult to handle and operate with such a voluminous data, and that creates an easy path for the criminals to gain the benefits from the common people. Digital crime is growing massively through the computers, hard drives, USBs, disks, networks and other hand held digital devices such as mobile phones, PDAs etc. [4]. The technocrat criminals use very smart and modernized techniques for their bad intention. According to the FBI in 2012, more than 8 billion of the different property loot was recovered [35]. Digital forensic is an alternative approach to overcome this subtle situation. Digital forensic is a discipline of forensic science that deals with digital data, methods for establishing the identity of a criminal through proper investigations. Several Computational Intelligence methods (fuzzy method, artificial intelligence, rough sets, genetic algorithms etc.) and pattern recognition techniques are serving in this field. Forensics bears very old historical background, and it dates back to the ancient roman era. Digital forensic was introduced in 1980s.The advancement of digital forensics is not commensurate and sophisticated enough with the development of criminology. In this dynamic field, the literary contribution is not at a satisfactory level and hence it desperately requires requisite attentions. It is still believed that, this is a developing area of research seeking robust solution and technology to save our planet from the crime and terrorism. Biometric technology is a key constituent of modern forensic and surveillance technology. Data from different digital resources (image, audio and video) of related biometric modes are accumulated for scientific investigations [23]. Fingerprint and DNA are two well-known and matured techniques and other distinguished modes with different features and tools are utilized as evidences. Hand biometrics is one of such relevant modality. Here, the future prospective of different hand based modalities (e.g. hand geometry, palmprint, hand vein etc.) in forensic have been addressed. Among all hand based features available, palmprint carries the most significance and hand-bacteria identification is a new promising area [30]. Hand geometric designs are thoroughly studied, and its implications have been sorted out. Dorsal hand vein pattern and handprint are also reported.

This chapter is organized as follows: Section 2 describes the biometrics and multibiometrics. Section 3 is presented with digital forensic techniques and the scope of biometrics in digital forensics. Section 4 is contained with the details about various hand based biometric modes for forensics. Finally, conclusion is drawn in Section 5.

## 2   Biometrics

'Biometric' refers to the different physiological (e.g. face, fingerprint, iris, retina, DNA, hand geometry, etc.) and behavioral (e.g. voice, gait, signature, keystroke etc.) uniqueness of a person acquired from different human organs. It is an automated pattern recognition system that allows access grant only to the enrolled users. These inherent human properties can't be easily stolen, spoofed or shared by third party, and users need not memorize them. Thus, biometric is considered as the best substitution of conventional knowledge based (password) and token based (ID card) secured system where possibility of identity thievery is high. The foremost job of a biometric system is to discriminate whether a claimed identity is legitimate or not. The major properties of a biometric system are uniqueness, universality, stability, measurability, acceptability and performance [1]. Some distinct properties from any particular mode of an individual are extracted and stored in the database as feature templates. Feature vectors are compared against the stored templates, and depending on a predefined threshold value matching is performed. The matching score determines whether the claimant is a valid or unauthorized user. A user can be tested for identification (one-to-many) or verification (one-to-one) depending on the application necessity by applying different classification algo-rithms (e.g. Artificial Neural Network (ANN), K-Nearest Neighbors (KNN), Support Vector Machine (SVM) etc.). The performance of a biometric system is measured by the following parameters: False Accept Rate (FAR), False Reject Rate (FRR) and Equal Error Rate (ERR). The parameters are plotted in Receiver Operating Characteristic (ROC) curve. The FRR and FAR should be minimum in forensic and highly secured environment, respectively.

A biometric system works in mainly four modules: sensor, feature extraction, database and decision module [1, 18].

**Sensor module:** raw images from respective mode(s) are collected by the forensic experts using high quality cameras or scanners from the evidences at the place of crime occurred. Image resolutions for different modes are also been standardized (e.g. 500 ppi for palmprint). A set of different images are collected at different angle and pose. Noise could be associated with images and environmental factors may degrade the quality of images.

**Feature module:** after noise removal and image quality enhancement, certain uniform and unique features are extracted from the underlying trait(s) for commencing the research. Features are stored in either encrypted or latent form, to protect their identity from intruders.

**Database module:** feature database is stored in high capacity storage devices such as hard drive or disk. The database size varies according to template size of applied mode and number of enrolled users. This module is very sensitive to attack by the hackers or criminals directly or indirectly from computers or through networks.

**Decision module:** to find matching score with respect to a pre-specified threshold, test feature vector is compared with the feature vectors stored in the database. The absolute, Euclidean, Hamming or Mahalanobis distance functions are usually applied. Sometimes, the similarity score is expected to be normalized using min/max, median, z-score etc. based techniques [6]. Depending on the score, final decision is made to recognize a person as authentic or unauthorized person.

Biometric systems are developed using a single mode or several modes and categorized as unimodal or multimodal [1] system, respectively. A multimodal system uses certain types of fusion based techniques [6]. Some remarkable benefits over unimodal systems [18] include the following: (i) flexibility and universality (ii) improved matching accuracy (iii) "spoofing" attack minimization (iv) noise effects reduction (v) better reliability etc. Pre-matching fusion is applied at the sensor or feature extraction module and post-matching fusion is applied at the matching and decision module [6]. Fusion is applicable at every level: sensor level (2D and 3D imaging), feature level (similar features for the same trait like hand geometry, dissimilar features for different modes such as hand geometry and palmprint), rank level (some best results are arranged in an order), score level (matching scores are combined or normalized) and decision level (two or more decisions from different sources are combined). Decision level and score level [11, 28] are two general fusion methods whereas sensor level and feature level [13, 27] are getting more observations. Related contributions on some fusion based methods are described in Section 4.

## 3   Digital Forensics and Biometrics

Digital forensic is derived from computer forensics. It can be defined as the specialized and scientifically proven methodologies, used for the reconstruction of events to identify criminal or unauthorized actions. Forensic Research Workshop (DFRWS) Technical Committee has defined digital forensic science [4] as: "The use of scientifically derived and proven methods toward the preservation, collection, validation, identification, analysis, interpretation, documentation and presentation of digital evidence derived from digital sources for the purpose of facilitating or furthering the reconstruction of events found to be criminal, or helping to anticipate unauthorized actions shown to be disruptive to planned operations."

The necessary steps of digital investigations are collection, preservation, examination and analysis. Different models of forensics are available and in [7] some well known models (DFRWS (2001), EDIP (2004), CFFPTM (2006) etc.) are discussed. The major challenges of this domain include data encryption, anti-forensics, wireless technology, data size and many others [9].

**Collection:** it is a critical step, involves finding out and collecting the information and evidences (biometric evidences and e-evidences) relevant to the research. Finally, data is stored in digital format.

**Preservation:** keeping the collected information safely so that no damage can be done during the process.

**Examination:** scientific and systemic process of research, also known as "in depth systematic search of evidence" related to the event.

**Analysis:** based on the previous step decisions are taken about the event.

The last two steps are time consuming and depending on the importance of the crime. The entire research process is solely dependent on the group of proper evidences. The Generic Computer Forensic Investigation Model (GCFIM) is described in [7], including two additional phases (pre-process and post-process) along with these general phases. Digital forensic is differentiated into mainly four different categories: computer, database, network and mobile forensics.

**Computer:** it justifies with the present state of a digital system such as computer, browsing history, the storage medium, last accessed and log files etc.

**Database:** contains the information about database, metadata and log files.

**Network:** analysis of a networked system (LAN/WAN or internet connectivity) are performed by observing the network traffic, delay and data transfer.

**Mobile:** it handles with the call details, SMS, MMS, browsing history and video clips of the device. Smartphones and 3G enriched with more facilities compared to ordinary cellphones can be used for retrieval of related information. Through the GPS, finding of locations is possible. Crimes through internet and mobile devices have increased unexpectedly, and it surpasses every year.

Biometric system works with digital images (2D or 3D), audio and videos. Each of these medium carries their own features and utility. Basically, the strengths of biometrics are employed in creating a strong forensic tool. Forensic is a post-event and biometric is a pre-event phenomenon. In user authentication, one or more particular mode(s) are predefined to access a secured environment. The system is user friendly, and users are cooperative. The reverse situation exists in forensics; the mode(s) is (are) not predefined. It is determined depending on the collectability of evidences associated to the crime. Event reconstruction is the principal challenge for the investigator. Biometric is a real time authentication system and checks liveliness of a user and processing time is low. But, in forensic liveliness is not an important factor, and it is time consuming to reach a final decision.

Biometric is universally implemented in different domains of government, commercial and forensics. For higher security environment retina, facial thermograms, iris, fingerprints are commonly used. Though, the gait is not much exploited, but it is important for real time surveillance systems, where a criminal or suspicious person can be traced by the CCTV footage. Several cases have been solved using this technology. Whereas the other competitor modes: latent fingerprint, palmprint, footprint, voice, DNA profile, and dental radiographs are obtained according to availability related to the event, are utilized for forensic investigations as evidence to identify the criminal. From 1980s, fingerprint bears its responsibility for the FBI [35]. In 1988, FBI introduced DNA analysis for research. After 10 years, in 1998 NDIS (National DNA Index System) was developed as part of CODIS at national level (USA) containing the DNA profiles [35]. Both are dominating other biometric modes since their origin. But, the main problems are the collectability and higher implementation cost because these biometric evidences cannot be collected from all the events.

In biometric applications, the size of the population is either low or medium. Practically, in forensic investigation user search space is millions which is one of the most important factors for any biometric system to provide maximum accuracy in such an extensive search space. From the evolution of biometric to present a situation, unfortunately, a particular method does not exist which can be described as the 'best' in all kind of applications. Not all of these modalities are examined properly. Other than fingerprint and DNA analysis, face and palmprint are already exploited in forensics. Hand geometry, footprint, voice and gait recognition are these exploratory areas that need to be more focused. The Biometric Centre of Excellence (BCOE) also wants to improve the potential in the near future, and their next generation approach will be "bigger, faster, and better" than the Integrated Automated Fingerprint Identification System (IAFIS) [35].

## 4 Hand Based Biometrics

The hand is an important interface of human to perform most of the works in our daily life. It is one of the oldest biometric features used for authentication [8, 34]. It is not yet explored in criminal investigation and law enforcement. It may be useful in certain circumstances where other modalities (e.g. DNA, face etc.) are unavailable or unreliable. Other significant benefits it provides are reliability, convenience, non-intrusive, user friendliness and immutable as it is dependent on the intrinsic physical properties directly or behavior of a person. A number of distinct features are measured from different parts of either side of the hand and used as biometric properties. From the front side of the hand, features of palmprint [3, 12], handprint (especially for infants) [29], hand body [13], finger geometry (specific fingers) and hand geometry are measured [34]. From the back side of the hand, vein pattern [2, 17], dorsal hand form [20, 25] and finger knuckle print [10, 16] are commonly considered as biometric characteristics and named according to part of the hand associated with this purpose. All the related hand based modes are shown in Fig.1.

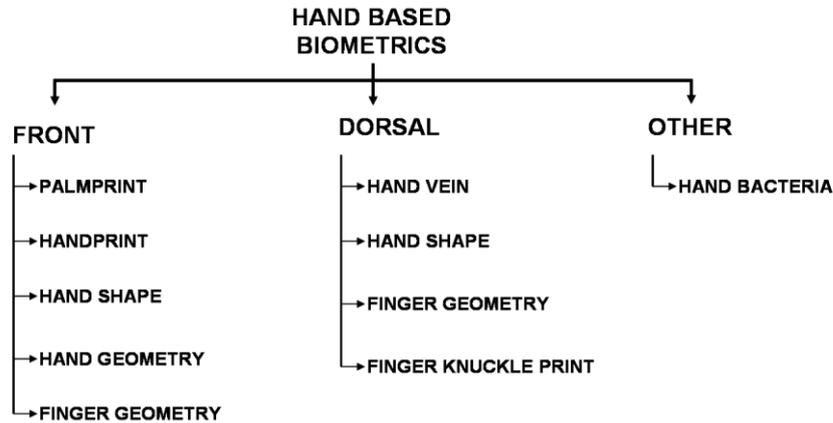

**Fig. 1.** Classification of hand based biometrics.

Palmprint and hand geometry are two general and conventional hand based recognition systems. Performance of hand-shape (frontal or dorsal) detection is satisfactory comparatively when fusion based techniques are implemented. Dorsal vein pattern and finger knuckle prints are emerging the field of hand biometric system. But, using these modes no business exists for digital research purposes. Palmprint produces very high accuracy as compared to fingerprint. It is considered as a better substitute of the other because more discriminative features can be extracted as larger surface area of the palm. Hand vein recognition is rendering very significant performance. Another pertinent field is hand bacteria identification. Though it is not regarded as biometric directly, but it resides in hand surface, producing excellent correctness.

### 4.1    Hand Geometry

Hand geometry mainly includes the measurement of many consistent geometric features of fingers (such as length and width at different positions of fingers) and hand form (such as palm area and palm width). *'Identimat'* was the first successful hand biometric system used for person authentication since 1980s. Afterwards, other devices were developed using this modality in commercial and government applications such as attendance maintenance, Olympics (introduced in 1984), nuclear plants and many others [8]. The number of different geometric features generally lies within the range of 20 to 40. Feature template size is significantly low, requiring only within 10 bytes and having lesser processing time (about 5 sec.). Other important benefits are easier to access, and environmental parameters such as bad weather, dry skin and lighting conditions can't alter system performance heavily. But, lack of high uniqueness of hand features as compared to fingerprint features or DNA profile is a major concern for adaptation in forensics. It is inapplicable for identification because of the larger population search space and thus it suffers from the scalability problem. In the verification, it's a suitable alternative. Most of the unimodal system allows either the left or right hand. Whereas some fusion based systems are developed using both hands [8].

In early days of hand biometrics, a CCD camera or scanner was used for image acquisition, and the image quality was very low (less than 100 ppi). The pose of hand placement was fixed by using 'peg'. A rigid pose minimizes the inter-class and intra-class pose variations, finger alignments and maintains uniform spacing between fingers. This imaging system reduces misclassification rate. The major problem is larger device dimension and impossible to embed in smaller devices (laptops) and may causes hygienic issues for personal. Recently, high resolution digital camera and infrared (IR) camera [19, 26] are applied that facilitates without any pose restrictions, providing more user flexibility. These imaging systems are lesser error prone and lesser noise sensitive, but cost of the device is greater. Modern research interest on hand geometry is paying attention to 3D images and its fusion with 2D images as well [15]. Data fusion is performed with the other hand related modes such as palmprint, finger knuckle print or hand vein model for robust and reliable solution [21, 27]. Multibiometrics employing fingerprint and/or palmprint along with hand geometry is cost effective due to single working sensor [32]. Other than these modes, human face is considered as other important fusion mode [14]. The fusion techniques are applied at various levels. Brief studies of some general works are given in Table 1.

Some major complexity arises in hand geometry are:

i) The most challenging issue is to create a standard orientation of all images at pre-processing stage, before feature calculations. Freeness of hand placement causes angle variations between fingers and major axis. Incorrect finger alignment can't locate finger tip and valley points accurately.

ii) Any hand gadget; ornament or bracelet can affect wrong hand contour or form and determination and feature calculation. Stylish fingernail especially for women can play an important role for the same.

iii) Any damage or injuries in hand can prevent the usage of this mode.

iv) Hand shape changes over time and age. Silhouette at childhood is changed at different ages of a lifespan.

It is appropriate for low or medium security based applications with moderate population size. Fusion based hand biometrics is more interesting than single mode. All the related systems support the real time authentication.

**Table 1.** Some state-of-the-arts methods of Hand Geometry

| Author | Mode | Classification technique | Image quality | Database size | Accuracy |
|---|---|---|---|---|---|
| [5] | 2D | i) Hausdorff distance of hand contours, ii) independent component features (ICA1 and ICA2) of hand silhouette images. | HP Scanjet 5300c 383× 526 pixels at 45 dpi. | 1374 right hand images from 458 subjects. | Verification: i)Hausdorff: 97.36%. ii) ICA1: 97.2%. ICA2:98.2% |
| [8] | 2D | Independent Component Analysis (ICA) on global hand appearance of both hands. | Flatbed scanners, at 150 dpi. | 918 subjects, 3 images per left & right hand per user. | Verification: Left: 1% EER. Right: 1.16% EER. |
| [11] | Eigen palm and Eigen finger. | Feature extraction by Karhunen-Loeve (K-L) transform. Final decision by the modified k-NN. | Low-cost scanner at 180dpi. | 1,820 hand images of 237 people. | Identification: 0.58% EER. |
| [13] | Hand shape and palm texture. | Correlation-based feature selection (CFS) algorithm. KNN Classifier by minimum Euclidean distance. | Digital camera 300×300 pixel. | 1000 images of 100 subjects, 10 images per subject. | Recognition: 97.8% |
| [15] | 2D & 3D hand geometry and palmprint. | 3D palmprint, represented by SurfaceCode. Matching score level fusion. | 3D digitizer. 640×480 pixels. | 3540 right hand images from 177 subjects. | ERR: 2.3%. AUC (area under the ROC curve) : 0.9888 |
| [19] | 2D thermal images. | Different geometric features of a hand are extracted. Recognized by Extension theory. | Infrared camera. | 300 images, 30 persons, 10 images per user. | Accuracy: 92% |
| [20] | 2D Dorsal hand geometry + fingerprint. | Min-max scores normalization. Distance based matching with respect to the threshold. Feature level (same mode) and score level (multimode). | NIR camera. 240×320 pixels, at72 dpi. Veridicom sensor for fingerprint, at 500 dpi 300 ×300 pixels. | 100 users, 5 images for each mode of left and right hand. Total 30 images per users. | EER: 0.0034% |

### 4.1.1 System Model

A simple plan, which is discussed in Fig.2, is to find out the suitability of hand biometrics in the forensic domain. In traditional hand biometric system, hand type (left/right) and number of sample images (for enrolment and testing) are predefined which is one of the most unavoidable limitations. Although, there may some accident or fracture on the hand that has been used for enrolment phase cannot be altered at some later time. But, in case of forensic the type of hand can not be pre-specified. It is defined according to the collected evidences. So, in such a situation, this present scheme is suggested which can determine the hand type automatically, and features can be extracted accordingly. A robust technique is needed for calculating accurate features. So, the most significant step is hand image normalization. It consists of several sub-steps which are environment elimination, rotation, irregularities removal at wrist region, determination of hand type and finger tips and valleys localization. Many features are calculated from the normalized hand images. Finally, users are classified as genuine or imposter according to the classification algorithm.

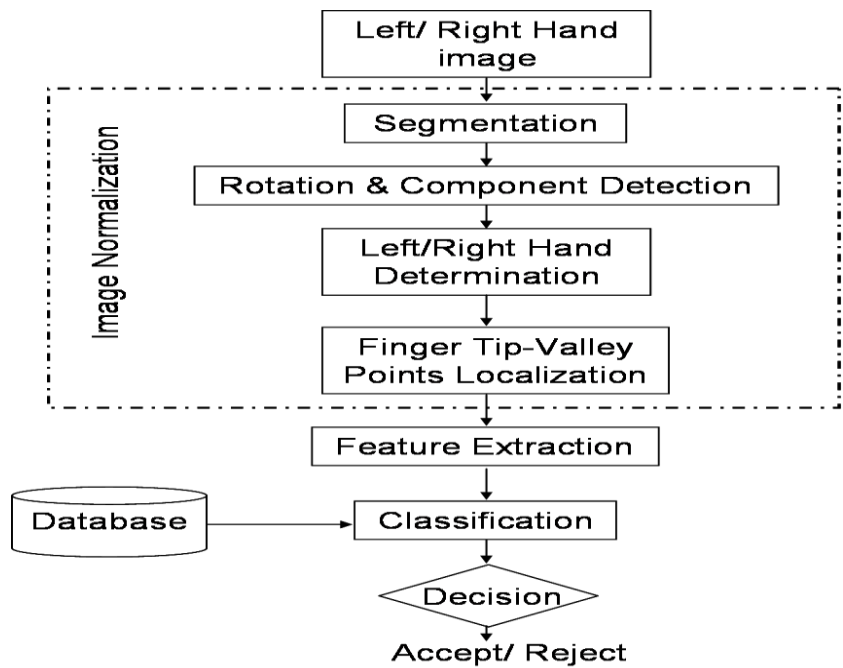

**Fig. 2.** Hand geometry based forensic system model.

*Hand Image Segmentation.*

Collected raw images consist of the actual hand texture with or without any stuff as foreground and a dark background. The first step is removal of undesirable background and noise associated with images. Conversion from an original color image into a grayscale image (Fig.3a) is performed using thresholding by the *Otsu's* method, and median filter is applied for noise removal. Images are then converted into binary (Fig.3b) from the grayscale images and hand contour is detected using *'Sobel's* edge detection method, as shown in Fig.3c.

*Rotation and Component Detection.*

The hand images are acquired at different posture, different angle with their hand gadgets; only with a little attention that no two fingers should be attached or overlapped. So, to define a standard uniform template, a particular orientation for all the images is employed by rotating at various angles (90,180 degrees) as required to make them perpendicular with the major X-axis. Due to any hand ornaments or wristwatch used by a person during image acquisition, the binary image may contain several components. The largest part is detected, and remaining small components are ignored. A reference line (AB) at certain distance, above from the bottom of that part is considered by scanning the pixels from the left to the right direction. The leftmost and rightmost nonzero pixels are the two end points of the reference line. The lower portion of line AB is neglected, and its midpoint (R) is defined as the reference point (Fig.3d).

*Left or Right Hand Determination.*

To decide whether the given hand is the left hand or the right hand the given input is needed to locate the tip region of the thumb. Generally, for left or right hand thumb tip region, the leftmost or the rightmost nonzero pixel above reference line is traced, respectively. Then, the reverse extreme pixel in the opposite direction for either hand is considered. To check whether the image is of left or right hand, examine these two extreme pixels. Say, the leftmost pixel is LM, and the rightmost pixel is RM. If the leftmost pixel LM is below the rightmost pixel RM with respect to the R, then the finger is thumb (related to LM), and other is little finger (related to RM) and that hand is considered as a left hand. Similarly, reverse reason is followed for the opposite hand. This identification of hand type is helpful for finding out the positions of finger tips and valleys because the space and flexibility between thumb and index finger is not same as between ring and little finger.

*Locating Finger Tips and Valleys.*

A general algorithm has been applied to both the hands to locate tip and valley points and based on which features are extracted. Once the thumb tip region is located, little finger tip region can also be found out in a similar manner. From those tip regions, exact tip features can be traced on hand contour by scanning the pixels. Other fingertip positions are located using the maximum Euclidean distance from the reference point. First of all, middle finger tip is identified, and then tips of index and ring finger are placed. The valley points between any two fingers of either hand are determined by scanning the hand contour. Other valley points of the thumb, index finger and little finger are defined using equal Euclidean distance from the tip to the other valley point, placed at the opposite side of that particular finger. All the key points are marked (by point 1, 2, 3 etc.) in Fig. 3f.

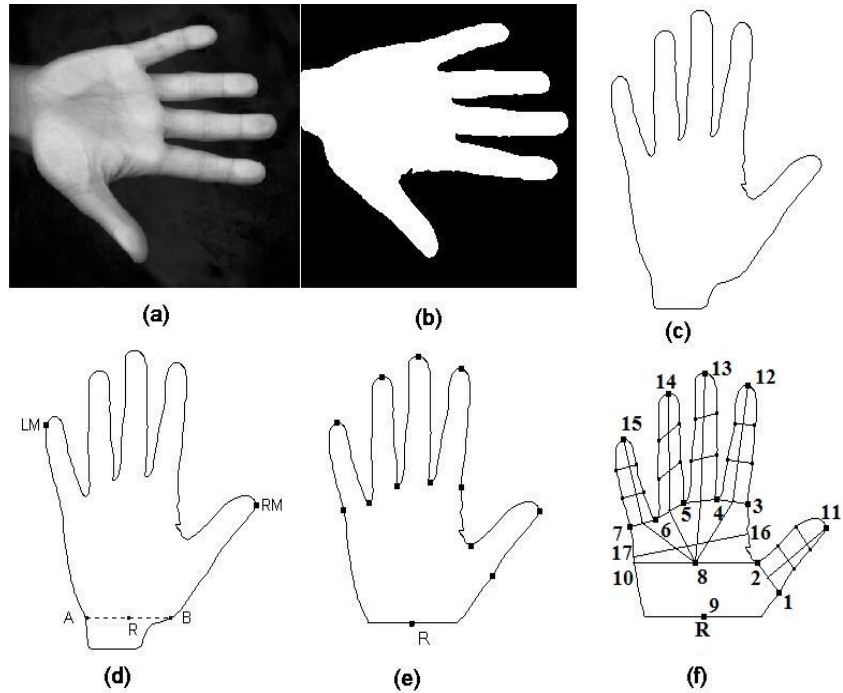

**Fig. 3.** (a) Grayscale hand image, (b) binary image, (c) rotation and hand contour, (d) determination of the right hand and guillotining, (e) locating landmarks points, (f) feature extraction.

*Feature Extraction.*

After locating the landmark points, some unique features from any normalized hand image are extracted by measuring lengths and widths at one-third and two-third position of the finger length of individual fingers. The widths of palm at two different locations and distance from the midpoint of palm width line to the middle of every finger baseline, except the thumb are computed. Total 26 hand features are measured from each normalized image. Most of the features are widely accepted in many previous works, and some new features are incorporated.

The feature set is defined as follows: 5 finger length (1 per finger); 10 finger width (at 1/3 and 2/3 position of every finger); 5 finger baseline width (1 per finger); 2 palm width (at different position of palm; depicted by line 2-10 and 16-17) and 4 lengths from the middle point of palm line (illustrated in Fig. 3.f as midpoint 8 of line formed between point 2 and 10) to mid of finger base-lines excluding the thumb. All features are graphically shown in Fig.3f, and the number of features can be increased.

*Classification.*

Persons are recognized by the minimum distance classification algorithm. At first, the Euclidean distance between the test feature vectors and the stored templates are calculated. Based on the minimum distant, summation of the features from a particular subject is considered as a recognized class. Consider, $A_{m \times n}$ is an enrolled feature matrix; where, row *m* represents extracted features for all subjects and column *n* is the feature vector for every enrolled image. $B_{k \times n}$ is the test feature matrix; where, *k* is used for testing for every subject and *n* is same as defined above. Matrix $C_{m \times n}$ is used to store the Euclidean distances between every test vector with all rows of 'A' matrix. Calculate the Sum (**S**) for every row of **C** matrix. If multiple samples are used during the enrolment phase then, average of Sum is computed to assign final class label. Formally, the algorithm is given below.

**Input:** Enrolled feature matrix(A), test feature matrix(B).
**Output:** Recognized Class label.

1. $C_{i,j} = \sqrt{(A_{i,j} - B_{x,j})^2}$
2. $S_i = \sum_{j=1}^{n} C_{i,j}$
3. Class $= min(Avg(S_i))$
4. End.

Where, $1 \leq i \leq m$, $1 \leq x \leq k$ and $1 \leq j \leq n$; A, B, C and S are defined above.

*Experimental Results.*

The database is collected from E. Yörük et al. [5]. In this work, the database contains images of 253 users, 3 images for each hand of a person. 157 left hand user and 96 right hand users are considered. First image with 383×526 pixels and finally at feature extraction step the images are resized to 200×300 pixels. The population is a blend of left and right hand subjects; the result is calculated with the distance threshold value of 3.2. For enrolment (size = K), one and then two images per subject are applied, and only one image is used for testing and the results are given in Table 2.

**Table 2.** Performance evaluation.

| Enrolment Size (K)=1 | | |
|---|---|---|
| **Hand Type** | **Minimum Threshold(t)** | **Recognition Rate(%)** |
| Left | 5.6 | 94.9 |
| Right | 5.8 | 96.88 |
| Combined | 5.8 | 95.65 |
| **Enrolment Size(K)=2** | | |
| **Hand Type** | **Minimum Threshold(t)** | **Recognition Rate(%)** |
| Left | 2.8 | 98.09 |
| Right | 2.4 | 100 |
| Combined | 2.8 | 98.81 |

With different population size, the recognition rate is also computed for K=1 and 2, shown in Table 3. The performance is not up to a satisfactory level for K=1, which is 3.16% lower than K=2. The experimental result shown in Table 3 is carried out by varying the group size from 50 to 253, at five steps.

Table 3. Performance with different population size.

| Population Size | 50 | 100 | 150 | 200 | 253 |
|---|---|---|---|---|---|
| Recognition rate (K=1) | 100 | 99 | 97.3 | 96.5 | 95.65 |
| Recognition rate (K=2) | 100 | 100 | 99.3 | 99.5 | 98.81 |

The experimental results mean that, the system provides acceptable performance for the medium security applications with improved 98.8% accuracy for 253 combined subjects.

### 4.1.2 Applications and Scope in Forensics

Hand geometry recognition systems are not widely used for authentication purpose. Approximately, 10% of systems are used for user verification in academic institutions and industrial purposes. Examples: Sacramento County, California, was using hand geometry. The INSPASS (The United States Immigration and Naturalization Service (INS) Passenger Accelerated Service) System used hand geometry for verification. Wells Fargo Bank uses hand geometry to prevent from unauthorized access to the bank's data centres. An elementary school in New Mexico has also presented hand geometry system. The University of Georgia has used a hand geometry system since 1972 to identify students on the unlimited meal plan, thus preventing them from lending their cards to others. In a lot of other domains, this trait is being used since the last century.

Images and videos of children engaged in sexual activities are increasing the number of criminal cases involving computers. An investigation by the RCFL (Regional Computer Forensic Laboratories) implied maximum percentage of child pornography. Hand biometrics is used in forensic cases such as child abuse, sex exploitation, kidnappings and missing infant identification [23]. Studies imply that in India more than 7,200 children, including infants, are raped every year, and still many cases go unreported [33]. About 53.22% children faced different forms of sexual abuse and in 83% of the cases parents were involved positively. Structure of hand carries the useful information of determining gender and age. So, if the hand images from such activities are collectable, then decision can be taken whether a child is involved or not in those crimes. The Forensic Audio, Video and Image Analysis Unit (FAVIAU) is also interested in this section [23].

It can be prevalent in certain situations from where any strong evidence is not accessible. Most of the criminals use facial masks and hand gloves and most of the time the event is pre planned. The situation can illustrate as criminal used mask and gloves very often during the crime from where it is very complicated to collect evi-dences and identify the person by face, fingerprint or DNA. Even if the gloves are damaged too. Form an unidentified body remains due to some accident or blast, the initial research can be started with. Same reason is also legitimate for deformed body identification. Another situation may take place due to environmental factors. Hand with dirt, oil, water, blood or any kind of liquid, from where DNA can't be retrieved, and reliability of fingerprint is minimized. So, hand geometry can be the best option to be utilized in such circumstances.

### 4.2  Palmprint

Palmprint is a well known biometric technology, introduced in India by Sir William Herschel in 1858. In 1994 palmprint supported system were developed by a Hungarian company. It carries similarities with fingerprint. It consists of the friction ridge information such as ridge flow, ridge structure, major palm lines etc. Palmprint is believed as a stronger mode with certain properties like, uniqueness, universality, stability and collectability for authentication. Palmprint offers many advantages over fingerprint and hand geometry [12]. Palmprint area is larger than the fingerprint area and thus carries more robust information than a fingerprint about the person. Palmprint of twins are different, and the palm lines also ensure genetic disorder of people. High resolution (at least 500 dpi) palmprint is suitable for forensic [3] and low resolution (at most 400 dpi) palmprint is useful for authentication. The features include the principal lines, ridges, wrinkles, minutiae and delta points of the palm. Some common subspace based methods used for this mode are Linear Discriminant Analysis (LDA), Principal Component Analysis (PCA) and Independent Component Analysis (ICA). Statistical and transformed based techniques are also developed. The system should support partial-to-full matching scheme [12]. The accuracy of palmprint is comparable with the fingerprint and DNA. Surveys from law enforcement resources imply about 30% evidences collected from a crime scene contains palmprint [3]. But, main difficulty is collectability. Images of palmprint from different objects are collected with noise or overlapping which create complications in feature extraction. The most difficult job is correctness improvement of partial-to-full matching. Rotation of palmprint at any random angles is another key factor. Minutiae-based latent-to-full matching by the *MinutiaCode* is described in [12], and the performance is evaluated with live-scan partial palmprint and latent palmprint against background samples of full palmprint. Matching is performed both locally and globally using improved feature selection methods. Rank level fusion of latent palmprint is experimented using OR rule to test whether two latent palmprint represents the same or a different person. The radial triangulation based approach is proposed by [22] with full palmprint and latent palmprint. The details of experiments are given in Table 4.

Handprint is another method that is closely similar to the palmprint. It includes all the printable area of the hand, including the fingers whereas palmprint only includes the palm area. Handprint is mainly suitable for child identification as because their palm features are not different, reliable and invasive. On average, the hand area of infants is 2.5 to 3 times smaller than the adults. It means that the feature extraction is difficult because baby hand features are very fragile and change over time. The image quality should be very high for better ridge extraction using this mode. According to [29], it should be 1500 ppi and in their contribution images are collected at 1000 ppi. From the handprints, gender and height can be guessed and from the skeletal bone structure of the hand, age of can be estimated also. But, baby handprint can't be collected sometimes due to hygienic problems as their skins are very sensitive and oily. It has also been observed baby hands are closed most of the time. So, in such conditions footprints [24] are dependable.

Below16-24 ages people, are requiring better protection both in the real and cyber world. Newborn babies are being stolen from the hospital in most of the cities. Such a case happened in Mumbai, which leaded the Mumbai High Court to advise that all newborn babies should have their feet and handprint captured within two hours of birth. In developing nations like ours or Brazil, rate of child missing or swapping from hospitals is also high [29]. Face recognition technology could not be applied reliably below the age of twelve, and similarly voice recognition can't be used due to the undifferentiated nature of voice between boys and girls. So, newborn baby identification is not explored properly, and it needs a robust, low cost and faster solution in real time applications.

### 4.3 Hand Vein

Hand vein pattern recognition is one of the new research area that find it's suitability in criminal identification since the killing of the US journal reporter D. Pearl by the mastermind of 9/11 attack, Khalid Mohammed. This field is not matured enough through worldwide in terms of application domains. In 1997, Hitachi developed first vein pattern matching device and used in Japan and Korea for verification vastly. In Japan, approximately 80% banks use contact free finger vein biometrics. It is highly unique even for twins. It is more reliable and invasive than the other hand biometric systems. Study implies its performance in a constrained environment is very high (99.99%) comparable with DNA matching. Vein pattern includes the inner blood vessels, visible from the outer skin surface. It can't be manipulated or replicated externally, and it's very difficult for 'spoofing' attack, as well. No weather condition can hamper its performance. It is stable over time for adults. But for a child and older people the pattern changes over time. The vascular pattern is complex and different enough for pattern recognition. An infrared (IR) camera (mainly, Near IR) is employed for image acquisition (wavelength 800-1000 nm) from the back side of the palm. During preprocessing, noise is removed, and image quality is improved before feature definition. As it contains complex vascular pattern, some difficulty arises for background removal. Selecting the Region of Interest (ROI) is most significant challenge for feature extraction.

Finally, the feature vectors are matched against the enrolled database. Some general matching algorithms are Minutiae-based; Hausdorff or Euclidean distance based, correlation based etc. vein pattern matching is relevant in higher security based and forensics applications. Liveliness of a person can be verified in a contact free nature, makes it attractive. But, still no strong database presents of this mode and no forensic use have yet been developed.

**Table 4.** Study of some related hand based modes for forensics.

| Author | Mode | Classification | Image quality | Database size | Accuracy |
|---|---|---|---|---|---|
| [22] | Palmprint | Latent-to-full matching using radial triangulation method. | Full palmprint: 2304 x 2304. Latent print: 500 ppi. | 22 latent print, 8680 full palmprint form 4340 subjects of left and right hand. | Identification: (Rank-1):62%. |
| [12] | Palmprint | Minutiae-based latent-to-full palm-print. | Full palmprint: 1000 ppi. Latent print: 500 ppi. | i) 150 live-scan partial print ii) 100 latent palmprint, with 10200 full palmprint from Noblis and Michigan State Police (MSP) and Michigan State University(MSU). | Identification: (Rank1) i)78.9%, ii) 69%. |
| [3] | Palmprint | Fourier Transform of images and combines Modified Phase-Only Correlation with Fourier-Mellin Transform. | THUPALMLA-B and PV-TESTPARTIAL: both at 500ppi. | i) THUPALMLAB: 152 palms (1216 full and 1216 partial palmprint). ii)PV-TEST-PARTIAL: 10 users, 80 full and 40 partial palmprint. | EER: i) 27.1%, ii) 23.8%. With 101×101 inside lobe. |
| [29] | Handprint | Combined two stage approaches of Simulated Annealing and Oriented Texture Field (Finger Code-FC). | CrossMatch LSCAN 1000P sensor at 1000 ppi. (4964 ×5120 pixels). | Original Database (NB_ID): 1221 palm-print from 250 newborn at the University Hospital (Universidade Federal do Parana). 60 images form 20 newborn are experimented. | SA: At 0% FAR, GAR is 78%. FC: Rank 5. |
| [2] | Hand Vein | Multimodal Fat Distance (FD) + Max-Min Distance (MMD) SVM. | NIR camera set-up able to acquire an image of 320×240 pixels. | 342 sample fingers of 114 users, 3 samples per finger. | Identification: GAR: 94%, FAR: 0.15. |
| [17] | Hand Vein | Euclidean Distance based matching method. | Digital SLR camera with infrared filter and night vision lamp (940nm). | Database: (IITK) 1750 sample images of 341users. Absorption based method for image acquisition. | Verification: 99.26% at 0.03%.FRR. |

### 4.4 Hand Bacteria

A new direction in forensic investigation has been found by using the hand bacteria that live on the human hand surface which is 70-90% accurate, and its performance will be enhanced over time [30, 31]. The precision of this technology will be same as DNA or fingerprint. When blood, hair, fingerprint, saliva, palmprint or any strong evidence is not available from a complex and unfavorable conditions to determine the identity of the criminal, then this method will be most appropriate in the near future. Bacterial diversity of woman is higher than man [31]. Hand bacteria can be collected from the touched surfaces of the keyboards, mice or any object used for daily purposes and applied for the investigations. The bacterial DNA structure of the owner is more perfect and stable than any person even for twins. Bacterial communities on a finger, palm, fingertips or keyboard are distinct and very closely related to a person than others. About 150 bacteria species can be found, and they persist as long as 2 weeks at room temperature out of which only 13% of the species are shared between any two persons. The stability is very high, and the bacterial communities can be recovered even after washing out hands after hours. So, it would be simpler to accumulate evidences from the touched objects or surfaces using the bacterial DNA rather than DNA of that person.

Though the hand bacteria are not directly related to the traditional hand biometrics, but it could be a fantastic alternative tool for the next generation forensic identification. Researches in this excellent field are going on. A specialized and benchmark forensic tool can be developed in the coming years.

## 5  Conclusion

Hand based biometrics is serving many commercial and government domains throughout many decades. Apart from conventional user authentication, it can play a significant role in forensics. Different hand based modes carry some potential benefits which have been enlightened in this domain. Palmprint and dorsal vein structure are comparable to the DNA and fingerprint analysis in terms of accuracy, time and cost. Palmprint and handprint have already established their positions in forensics. According to our exploration, no work persists on hand geometry, finger knuckle print in forensics. Hand vein pattern is trying to find out its relevance in a criminal investigation. Hand bacteria identification belongs to the same category with enormous future possibility. Applications of all related traits are illustrated, and their scopes are depicted. The prime objective of related studies on these traits is to focus on their forensic aspects. Several multimodal fusion based approach are developed to perform well. A straightforward hand geometry recognition method is suggested for the same purpose. In next generation forensic, these hand modes can be utilized by the standard organizations, like FBI. So, we could expect from the forensic professional a strong forensic framework and tool using hand biometrics will be contributed. Finally, some research attention needed in this field in the hope that the applicability of hand based biometrics in forensic investigation will be improved in the near future.